\documentclass{article}
 \usepackage{amsmath}
 \allowdisplaybreaks[4]
    \usepackage{amsthm}
    \usepackage{amssymb}
    \usepackage{latexsym}
\usepackage{longtable,booktabs}
\usepackage{graphicx}
\usepackage{caption}
\usepackage{mathrsfs}
 \usepackage{cite}
\usepackage{color}
\usepackage{psfragx}
\begin{document}
\newtheorem{Lemma}{Lemma}[section]
\newtheorem{Proposition}{Proposition}[section]
\newtheorem{Theorem}{Theorem}[section]
\newtheorem{Corollary}{Corollary}[section]
\newtheorem{Example}{Example}[section]
\newcommand{\wzr}[1]{\textcolor{blue}{#1}}
\baselineskip 0.25in
\title{\Large\bf MISO hierarchical inference engine satisfying the law of importation with aggregation functions}
\author{{Dechao Li}\thanks{Email:\ dch1831@163.com}
\qquad Qiannan Guo\\
  {\small School of Information and Engineering},\\
{\small Zhejiang Ocean University,
  Zhoushan,
  316000, China}\\
{\small  Key Laboratory of Oceanographic Big Data Mining and Application of}\\
{\small Zhejiang Province,  Zhoushan,  316022, China}}
\date{}
\maketitle
\begin{center}
\begin{minipage}{140mm}
\begin{picture}(1,1)
 \line(1,0){400}
\end{picture}

\centerline{\bf Abstract} \vskip 3mm {\qquad
Fuzzy inference engine, as one of the most important components of fuzzy systems, can obtain some meaningful outputs from fuzzy sets on input space and fuzzy rule base using fuzzy logic inference methods. In multi-input-single-output (MISO) fuzzy systems, in order to enhance the computational efficiency of fuzzy inference engine, this paper aims mainly to investigate three MISO fuzzy hierarchial inference engines employed fuzzy implications satisfying the law of importation with aggregation functions (LIA). We firstly find some aggregation functions for well-known fuzzy implications such that they satisfy LIA. The fuzzy implication satisfying LIA with respect to a given aggregation function is then characterized. Finally, three fuzzy hierarchical inference engines in MISO fuzzy systems are constructed according to aforementioned theoretical developments. Three examples are also provided to illustrate our theoretical arguments.}

\begin{picture}(1,1)
 \line(1,0){400}
\end{picture}
\end{minipage}
\end{center}
\vskip 4mm
\section{Introduction}
\subsection{Motivation}
 \qquad As fuzzy systems can transform human knowledge into a nonlinear mapping, they have been successfully utilized in control, expert system, signal processing, decision making and so on. A fuzzy system mainly consists of fuzzyifier, fuzzy rule base, fuzzy inference engine and defuzzifier \cite{Wang}. The fuzzy rule base, which constitutes a set of fuzzy IF-THEN rules, is the heart of a fuzzy system. Usually, the fuzzy IF-THEN rules in an MISO (single-input-single-output (SISO)) fuzzy system have the following form
\\(SISO) \ \   \ IF \ $x$\ is\ $D_{j}$\ THEN  \ $y$\ is \ $B_{j}\ (j=1,2,\cdots, n)$,
\\ (MISO) \ \ \ IF \ $x_{1}$\ is\ $D_{1j}$ \ AND $\cdots$\ AND \ $x_{m}$\ is\ $D_{mj}$\ THEN  \ $y$\ is \ $B_{j}\ (j=1,2,\cdots, n)$.
\\ Where $\mathbf{x}=
(x_1,x_2, \cdots, x_m)\in U=U_1\times U_2\times\cdots \times U_m$ and $y\in V$ are the input and output variables of
the fuzzy system, $D_{ij}\ (D_j)$ and $B_j$ are respective fuzzy sets on $U_i\ (i=1,2,\cdots, m)$ and $V$. From the fuzzy logical point of view, the fuzzy IF-THEN rules can be regarded as a series of  fuzzy relations on $U\times V$. And
 they are often specified by the fuzzy implications. This brings about that more fuzzy implications are studied in order to meet the various needs for fuzzy systems \cite{Baczynski,Baczynski1,Fodor,Grzegorzewski,Lima,Massanet,Yager}.

As another important component of the fuzzy system, the fuzzy inference engine can transform the fuzzy IF-THEN rules
and fuzzy sets on $U$ into a fuzzy set on $V$ by some fuzzy logical principles \cite{Wang}. Especially, the generalized modus ponens (GMP) is often utilized in case where the rule base consists of unique fuzzy IF-THEN rule. The GMP introduced by Zadeh, as an extension of modus ponens (MP) in the classical logic, can be indicated straightforwardly as follows \cite{Zadeh}:
$$\textmd{Premise\ 1:}\ \textmd{IF}\ x\ \textmd{is}\ D\ \textmd{THEN}\ y\ \textmd{is}\ B\vspace{-2mm}$$
$$\textmd{Premise\ 2:}\ x\ \textmd{is}\ D'\qquad\qquad\qquad\quad\ \vspace{-3mm}$$
\qquad\qquad\qquad\qquad\qquad\qquad\begin{picture}(1,2)
 \line(1,0){170}
\end{picture}\vspace{-3mm}
$$\textmd{Conclusion:}\qquad\quad\qquad\qquad y\ \textmd{is}\ B',$$
where $D$ and $D'$, $B$ and $B'$ are fuzzy sets on $U$ and $V$, respectively.

To obtain $B'$, the compositional rule of inference (CRI) method is presented by Zadeh in 1973 \cite{Zadeh}. After, the generalized CRI methods are discussed by many researchers. Unlike the CRI method, Pedrycz proposes another inference method based on the Bandler-Kohout subproduct (BKS) composition denoted by $B'=D'\circ_{\textmd{BKS}}R$\cite{Pedrycz}. In Pedrycz's method, Premise 1 was translated into a fuzzy relation $R$ using a fuzzy implication, the conclusion of GMP problem is then computed as
$$B'_{\textmd{BKS}}(y)=\bigwedge_{x\in U}I(D'(x),I(D(x), B(y))),$$
where $I$ is a fuzzy implication.

Notice that there are still some deficiencies in the CRI method \cite{Baldwin, Mizumoto,Turksen,Wang1}. To compensate these deficiencies, the similarity-based approximate reasoning (SBR) method and triple implication principle (TIP) are considered \cite{Mizumoto,Pei,Raha,Turksen,Wang1}. Moreover, some commonly  acknowledged axioms (also inferred as GMP rules) are presented by Magrez and Smets in order to measure the availability of these inference methods for the GMP problem\cite{Magrez}.

Similarly, some standards should be required in order to assess the goodness of fuzzy inference engine. Considering the wide applications of fuzzy inference engine, the computational efficiency is one of the important standards for fuzzy inference engine. However, the computational complexity is a main drawback of the CRI method, Pedrycz's method and TIP method\cite{Cornelis,Stepnicka}. In order to overcome this shortcoming, employed the t-norms and fuzzy implications satisfying the law of importation, Jayaram studied a hierarchical CRI fuzzy inference engine \cite{Jayaram}. And it is shown that this hierarchical CRI inference engine is equivalent to the classical CRI inference engine. This implies that the MISO fuzzy system using the hierarchical CRI inference engine can be transformed into an SISO hierarchical fuzzy system.  It is worth mentioning that the fuzzy system with the hierarchical CRI inference engine has the advantage of computational efficiency\cite{Jayaram}. After, Stepnicka and Jayaram suggested another hierarchical inferencing scheme based on Bandler-Kohout subproduct\cite{Stepnicka}. Clearly, the law of importation plays an important role in these hierarchical inference engines. This inspires people to investigate the fuzzy implications satisfying the law of importation with t-norms and uninorms, respectively \cite{Baczynski2,Mas1,Massanet3,Massanet4,Massanet5}.

It is well known that a fuzzy IF-THEN rule is a conditional statement expressed by fuzzy propositions. The words are usually used to describe the fuzzy propositions. These words not only include those for describing attributes,
such as ``low", ``medium", ``high", etc., but also include those connectives for
connecting multiple propositions, such as ``not", ``and" and ``or"\cite{Zimmer1}. The t-norms and t-conorms have been applied to interpret the words ``and" and ``or", respectively\cite{Baczynski,Klement,Klir,Wang}. However, due to the vagueness of natural language and the subtleness of people's thinking, the connectives cannot be precisely modeled by a single t-norm or t-conorm under all circumstances, and sometimes they even do not correspond to any t-norm or t-conorm\cite{Zimmer1}. Indeed, the operators extended from t-norms and t-conorms can flexibly effectively handle the uncertainties in connectives that cannot be handled by t-norms or t-conorms\cite{Hudec, Wu}. For instance, the operators which are not necessarily required the associativity or commutativity have been used to model the words ``and" and ``or" in decision making and classification problems\cite{Bustince,Fodor}. Moreover, as de Soto et al. pointed out, a fuzzy mathematical model does not always have to be symmetric\cite{de}. This motivates people to seek some suitable operators describing the connectives. Aggregation functions, as the generalization of t-norms and t-conorms (see Definitions 2.4 and 2.9), have been applied extensively in fuzzy logic, decision making and classification problems \cite{Bustince,Dimuro,Fodor,Helbin,Liz,Mas,Mas1,Ouyang,Pradera,Pradera1,Pradera2}. Clearly, aggregation functions become increasingly concerned substitute for the t-norms and t-conorms in the actual decision making and classification.

 In the case of a singleton MISO fuzzy system, if a t-norm is chosen to interpret the word ``and" while the fuzzy implication $I$ is employed to translate the fuzzy IF-THEN rule, the solution of CRI method $B'(y)\ (\equiv1)$ is unless or misleading\cite{Li1,Li2}. The reason for it that $T(a,0)\equiv0$ and $I(0,b)\equiv1$ hold for any t-norm $T$ and fuzzy implication $I$, respectively. This triggers us to model the word ``and" by an operator $O$ meeting the condition $O(a,0)>0$ ($O(0,a)>0$) for any $a\in[0,1]$. Thus, we will investigate the law of importation with aggregation functions corresponding with the actual needs. And we develop three hierarchical inference engines based on the fuzzy implications satisfying the law of importation. For this purpose the law of importation is firstly extended as follows:
 \\{\bf Definition 1.1}\cite{Pradera} Let $A$ be an aggregation function and $I$ a fuzzy implication. $I$ is said to satisfy the law of importation with an aggregation function $A$ if for all $x, y,z \in [0, 1]$,
\begin{equation}
I(A(x,y),z)=I(x,I(y,z)).\tag{LIA}
\end{equation}
\subsection{Contribution of this paper}
\qquad As the argument above, the aggregation functions and fuzzy implications satisfying the law of importation play a pivotal role in computational efficiency of a fuzzy inference engine. Moreover,  the variety options of aggregation functions and fuzzy implications result in the flexibility of fuzzy inference engines. Therefore, we mainly develop three hierarchical inference engines employed the aggregation functions and fuzzy implications satisfying LIA in this paper. We first investigate some properties of aggregation functions and fuzzy implications which satisfy LIA. And then we seek the aggregation functions for the well-known fuzzy implications such that they satisfy LIA. With such aggregation functions and fuzzy implications, three hierarchical inference engines in MISO fuzzy system are developed. In a word, the contributions of this paper include:

(1) To study the properties of aggregation functions and fuzzy implications satisfying LIA.

(2) To seek the aggregation functions for the well-known fuzzy implications such that they satisfy LIA.

(3) To characterize the fuzzy implications satisfying LIA with a given \wzr{associative} aggregation function.

(4) To construct three fuzzy  hierarchical inference engines employed aggregation functions and fuzzy implications satisfying LIA.

This paper is organized as follows. Section 2 recalls some basic concepts utilized in this paper. In Section 3,  we study the properties of aggregation functions and fuzzy implications when they satisfy LIA. Section 4 shows necessary and sufficient conditions for $(A,N)$-implication \wzr{generated by an associative disjunctor} and R-implications \wzr{generated by an associative and commutative aggregation function} satisfying LIA. In Section 5, some \wzr{associative} aggregation functions are constructed for $f$-implication, $g$-implication, QL-implication, probabilistic implication, probabilistic S-implication and $T$-power implication satisfying LIA with them, respectively.
Section 6 characterizes the fuzzy implication satisfying LIA with  a given \wzr{associative} aggregation function.
In Section 7, three MISO hierarchical inference engines based on fuzzy implications satisfying LIA are developed. Section 8 provides three examples to illustrate our proposed methods.
\section{Preliminaries}
\qquad This section will recall the definitions of fuzzy negation,  aggregation function and fuzzy implication and their properties
utilized in the remainder of this paper.
\subsection{Fuzzy negation,  aggregation function and fuzzy implication}
{\bf Definition 2.1} \cite{Lowen} A fuzzy negation $N$ is a mapping on [0,1] which satisfies

(N1) $N(0)=1, \ N(1)=0$,

(N2) $N(x)\geq N(y)\ \textmd{if}\ x\leq y,\ \forall\ x, y\in [0,1]$.

 A strict negation $N$ fulfills

(N3) $N$ is continuous,

(N4) $N(x)> N(y)\ \textmd{if}\ x<y$.

A fuzzy negation $N$ is strong if

(N5)  $N(N(x))=x,
\forall\ x\in [0,1]$.

Moreover, a fuzzy negation $N$ is said to be vanishing (non-vanishing) if $N(x)=0$ for some $x\neq1$ ($N(x)=0\Longleftrightarrow x=1$), and filling (non-filling) if $N(x) =1$ for some $x\neq0$ ($N(x)=1\Longleftrightarrow x=0$).
 \\{\bf Examples 2.2}\cite{Lowen}
 \begin{itemize}
   \item The standard fuzzy negation $N_c(x)=1-x$.
   \item The smallest and the greatest fuzzy negations

\qquad $N_\bot(x) = \left\{\begin{array}{ll}
               1 & x=0\\
               0& \textmd{otherwise}
             \end{array}\right.$ and\quad
$N_\top(x) = \left\{\begin{array}{ll}
               0 & x=1\\
               1& \textmd{otherwise}
             \end{array}\right.$.
   \item The natural negation of a fuzzy implication $I$ (see Definition 2.11) is defined by $N_I(x)=I(x,0)$.
 \end{itemize}
{\bf Lemma 2.3}\cite{Baczynski} Let the fuzzy negation $N$ be continuous. The mapping $\widetilde{N}$ defined by
$$\widetilde{N}(x)=\left\{\begin{array}{ll}
 N^{(-1)}(x), & x\in(0,1]\\
    1 & x = 0
    \end{array}\right.$$
 is a strict fuzzy negation, where $N^{(-1)}$ is the pseudo-inverse of $N$ given by $N^{(-1)}(x)=\sup\{y \in[0, 1]|N(y)> x\}
$ for all $x\in[0, 1]$. Moreover,

i. $\widetilde{N}^{(-1)}=N$;

ii. $N\circ \widetilde{N}^= id$;

iii. $\widetilde{N}\circ N|_{\textmd{Ran}(\widetilde{N})}= id|_{\textmd{Ran}(\widetilde{N})}$,
where $\textmd{Ran}(\widetilde{N})$ stands for the range of $\widetilde{N}$ and $N|_{\textmd{Ran}(\widetilde{N})}$ denotes the restriction of $N$ to $\textmd{Ran}(\widetilde{N})$.
\\{\bf Definition 2.4}\cite{Grabisch} An aggregation function is a mapping $A:[0,1]^2\rightarrow [0,1]$ which meets

(A1) Boundary conditions: $A(0, 0)=0$ and $A(1, 1)= 1$,

(A2) Non-decreasing in two variables, respectively.

 Suppose that $f$ is a binary function on [0,1] and $\varphi$ an automorphism on [0,1] (that is, an increasing bijection on [0,1]). Defining the function $f_\varphi(x,y)=\varphi^{-1}(f(\varphi(x),\varphi(y))$, it is called as the $\varphi$-conjugate of $f$. Obviously, the $\varphi$-conjugate of $A$, denoted by $A_\varphi$, is again an aggregation function. Especially, $A_N$ is known as the $N$-dual of $A$ chosen $\varphi$ as a strict negation $N$.
 \\{\bf Definition 2.5}\cite{Klement} Let $A_1$ and $A_2$ be two aggregation functions. We say $A_1\leq A_2$ if  $A_1(x,y)\leq A_2(x,y)$ holds for any $x,y\in [0,1]$.
\\{\bf Definition 2.6}\cite{Grabisch} $e\in[0,1]$ is a left (right) neutral element of the binary aggregation function $A$ if  $A(e, x)=x\ (A(x,e)=x)$ for any $x\in[0,1]$. Further, $e\in[0,1]$ is a neutral element of $A$ if $A(e, x)=A(x,e)=x$.
\\{\bf Definition 2.7}  \cite{Pradera} Let $A$ be an aggregation function.

i. $A$ is a conjunctor if $A(1, 0) =A(0, 1) =0$,

ii. $A$ is a disjunctor if $A(1, 0) =A(0, 1) =1$,

iii. $A$ has zero divisors if there exist $x, y\in (0, 1]$ such that  $A(x,y) = 0$,

iv. $A$ has one divisors if there exist $x, y \in [0, 1)$ such that  $A(x,y) =1$.
\\{\bf Definition 2.8}\cite{Pradera1} Let $A$ be a binary aggregation function and $N$ a fuzzy negation. We say that $A$ satisfies the law of excluded middle principle (LEM) with respect to $N$ if $A(N(x),x)=1$ holds for any $x\in[0,1]$. Obviously, $A$ is a disjunctor if it satisfies LEM.
\\{\bf Definition 2.9}\cite{Grabisch} Let $A$ be a binary aggregation function. We say that $A$ is

i. associative if $A(x,A(y,z))=A(A(x,y),z)$ for any $x,y,z\in[0,1]$,

ii. commutative if $A(x,y)=A(y,x)$ for any $x,y\in[0,1]$,

iii. a semi-copula if 1 is its neutral element,

iv. a t-norm if it is an associative and commutative semi-copula,

v.a t-conorm if it is the $N$-dual of a t-norm,

vi. a uninorm if it is associative, commutative and $e\in (0,1)$ is its neutral element,

vii. a copula if it is a semi-copula which $A(x_1,y_1)-A(x_1,y_2)-A(x_2,y_1)+A(x_2,y_2)\geq0$ holds for all $x_1\leq x_2$ and $y_1\leq y_2$.
\\{\bf Example 2.10}\cite{Grabisch,Pradera2} The following are some distinguished conjunctors:

\begin{itemize}
  \item The smallest conjunctor, $C\bot(x, y)=\left\{\begin{array}{ll}
                                               1 & x=y=1\\
                                               0&\textmd{otherwise}
                                             \end{array}\right.$;
  \item The greatest averaging conjunctor, $(C_{avg})_\top(x, y)=\left\{\begin{array}{ll}
                                               0 & x=0\ \textmd{or}\ y=0\\
                                               x\vee y&\textmd{otherwise}
                                             \end{array}\right.$;
  \item Representable aggregation functions, $A(x,y) = g^{-1}((g(x\wedge y)-g(N(x\vee y))\vee 0))$, where $g:[0, 1]\rightarrow [0, +\infty]$ is   continuous strictly increasing with $g(0)=0$ and $N$ is a strong negation;

 \item Weighted quasi-arithmetic mean (WQAM), $M_{\lambda,f}(x, y)= f^{-1}((1-\lambda)f(x)+\lambda f(y))$, where $f:[0, 1]\rightarrow [-\infty, +\infty]$ is continuous and strictly monotone with $f(0)=\pm\infty$ and $\lambda\in(0,1)$;

 \item TS-functions, $TS_{\lambda,f}(x,y) = f^{-1}((1-\lambda)f(T(x,y))+\lambda f(S(x,y)))$, where $T$ is a t-norm, $S$ is a t-conorm, $\lambda\in(0,1)$ and $f:[0, 1]\rightarrow [-\infty, +\infty]$ is continuous and strictly monotone with $f(0)=\pm\infty$.
\end{itemize}
{\bf Definition 2.11}\cite{Baczynski}  A fuzzy implication $I$ is a mapping $I: [0, 1]^2\rightarrow[0, 1]$ satisfying

(I1) Non-increasing in the first variable, i.e., $I(x,z)\geq I(y, z)$ if $x\leq y$,

(I2) Non-decreasing in the second variable, i.e., $I(x,y)\leq I(x, z)$ if $y\leq z$,

(I3) $I(0,0)=1$,

(I4) $I(1,1)=1$,

(I5) $I(1,0)=0$.

According to Definition 2.11, for a fuzzy implication the following facts can be directly obtained

(LB) Left boundary condition, $I(0, y)= 1, \forall\ y\in[0, 1]$,

(RB) Right boundary condition, $I(x, 1)=1, \forall\ x\in[0, 1]$.
\\{\bf Definition 2.12} \cite{Baczynski,Massanet4} We say that the fuzzy implication $I$ fulfills

(NP) Left neutrality property, $I(1, y)=y,  \forall\ y\in[0, 1]$,

(IP) Identity principle, $I(x, x) = 1,  \forall\ x\in[0, 1]$,

(EP) Exchange principle, $I(x, I(y, z)) = I(y, I(x, z)), \forall\ x, y, z\in[0, 1]$,

(CP(N)) Law of contraposition with a fuzzy negation $N$, $I(x, y) = I(N(y),N(x)), \forall\ x, y\in[0, 1]$,

(OP) Ordering property, $I(x, y) =
1\Longleftrightarrow x\leq y, \forall\ x, y\in[0, 1]$,

(OP$_\textmd{U}$) Counterpart of ordering property for uninorms, $I(x, y)\geq e\Longleftrightarrow x\leq y, \forall\ x, y\in[0, 1]$ with $e\in(0,1)$.
\\{\bf Definition 2.13}\cite{Pradera}  An $(A,N)$-implication $I_{A,N}$ is a mapping $I_{A,N}:[0,1]^2\rightarrow[0,1]$ defined by $$I_{A,N}(x,y)=A(N(x), y),$$
 where $A$ is a disjunctor and $N$ a fuzzy negation. Further, $I_{A,N}$ is called an $A$-implication if $N=N_c$.  Moreover, $I_{S,N}$ is a strong implication or S-implication if it is generated by a t-conorm $S$ and a strong negation $N$.
\\{\bf Theorem 2.14}\cite{Pradera} $I$ is a fuzzy implication if and only if $I$ is an $A$-implication, i.e. there exists a disjunctor $A$ such that
$I(x,y)=I_{A,N}(x,y)=A(1-x,y)$.
\\{\bf Definition 2.15} \cite{Ouyang}  A function $I_A : [0, 1]^2 \rightarrow [0, 1] $ is called an R-implication if $$I_A(x,y) = \sup\{ t \in [0, 1]\mid A(x, t) \leq y\}$$
is a fuzzy implication, where $A$ is an aggregation function.
\\{\bf Definition 2.16}\cite{Pradera} A function $I_{A_1,A_2} : [0, 1]^2 \rightarrow[0, 1] $ is called a QL-operation given by
$$I_{A_1,A_2}(x, y) = A_1(N(x), A_2(x, y)),$$
 where $A_1$, $A_2$ are two aggregation functions and $N$ a fuzzy negation. Especially, a QL-operation $I_{A_1,A_2}$ is called a QL-implication if it satisfies I1 and I3-I5.
 \\{\bf Definition 2.17}\cite{Yager} An $f$-implication $I_f$ is a mapping $I_f:[0,1]^2\rightarrow[0,1]$ defined as
$I_{f}(x,y)=f^{-1}(xf(y))$ with the understanding $0\cdot\infty=0$, where $f:[0,1]\rightarrow [0,+\infty]$ is a continuous and strict decreasing function with $f(1)=0$.
\\{\bf Definition 2.18}\cite{Yager} Let $g:[0,1]\rightarrow [0,+\infty]$ be a continuous and strict increasing  function with
$g(0)=0$. A $g$-implication $I_g$ generated by $g$ is a mapping $I_g:[0,1]^2\rightarrow[0,1]$ defined as
$I_{g}(x,y)=g^{(-1)}\left(\frac{g(y)}{x}\right)$ with the understanding $0\cdot \infty=\infty$, where $g^{(-1)}$ is\vspace{1mm} pseudoinverse of $g$ given by $g^{(-1)}(x)=\left\{\begin{array}{ll}
                                                                                    g^{-1}(x) & x\leq g(1) \\
                                                                                    1 & \textmd{otherwise}
                                                                                  \end{array}\right.$\vspace{2mm}.
\\{\bf Definition 2.19}\cite{Grzegorzewski}  A probabilistic implication $I_C$ generated by a copula $C$ is defined by\vspace{2mm} $I_C(x,y)=\left\{\begin{array}{ll}
                                                                                   \frac{C(x,y)}{x} & x>0 \vspace{1mm}\\
                                                                                    1 & \textmd{otherwise}
                                                                                  \end{array}\right.$
if it satisfies I1.\vspace{1mm}
\\{\bf Definition 2.20}\cite{Grzegorzewski} A probabilistic S-implication $\widetilde{I}_C$ is defined as
$\tilde{I}_C(x,y)=C(x,y)-x+1$, where $C$ be a copula.
\\{\bf Definition 2.21}\cite{Massanet} Let $T$ be a t-norm. A $T$-power implication is  a function $I^T:[0,1]^2\rightarrow [0, 1]$ given by $I^T(x, y) = \vee\{r\in [0, 1] | y^{(r)}
_T \geq x\}$ for all $x, y\in[0, 1]$, where $y^{(r)}
_T=\begin{matrix}_{r\ times}\vspace{-0.5mm}\\ T(\overbrace{y,y,...,y})\end{matrix}$.
\\{\bf Lemma 2.22}\cite{Massanet} Let $T$ be a continuous t-norm and $I^T$ its power implication.\vspace{1mm}

i. If $T=T_\textmd{M}$ is the minimum t-norm, then $I^{T_\textmd{M}}(x,y)=\left\{\begin{array}{cc}
                                                                                                 1 & x\leq y\\
                                                                                                 0& x>y
                                                                                               \end{array}\right.$;

ii. If $T$ is an Archimedean t-norm with additive generator $t$, i,e, there exists a continuous strictly deceasing function $t:[0,1]\rightarrow [0,+\infty]$ with $t(1)=0$, then $I^T(x, y) =\left\{\begin{array}{ll}
                                                                                                 1 & x\leq y\\
                                                                                                 \frac{t(x)}{t(y)}& x>y
                                                                                               \end{array}\right.$.
\subsection{Similarity based reasoning and triple implication method}
\qquad  Let $F(U)$ be the set of fuzzy sets on $U$. To solve the GMP problem,  the algorithm for similarity based reasoning presented by Raha et al. as follows \cite{Raha}:

Step 1. Combine premise 1 and calculate $R(D, B)$ by some appropriate translating rules (such as a t-norm).

Step 2. Calculate $S(D', D)$ combining $D'$ and $D$ using a similarity measure.

Step 3. Modify $R(D, B)$ with $S(D', D)$ in order to get $R(D, B|D')$ utilized some schemes.

Step 4. Obtain $B'$ as $$B'(y)=\mathop{\bigvee}\limits_{x\in U}R(D,B|D')(x,y).$$

To obtain $R(D, B|D')$, they also proposed the following three axioms:

(AX1) $R(D, B|D')(x, y)=R(D, B)(x, y)$ if $S(D', D)=1$;

(AX2) $R(D, B|D')(x, y)=1$ if $S(D', D)=0$;

(AX3) $R(D, B|D')\supseteq R(D, B)$ holds for any $D'\in F(U)$.

Then $R(D, B)$ is consider in the following ways:

Case 1. $R(D, B)(x, y)=T(D(x), B(y))$, where $T$ is a t-norm.

Case 2. $R(D, B)(x, y) =I(D(x), B(y))$, where $I$ is a fuzzy implication.

Finally, the conclusions $B'_{\textmd{SBR}}$ and $B''_{\textmd{SBR}}$ are obtained as
$$B'_{\textmd{SBR}}(y)=\mathop{\bigvee}\limits_{x\in U}I(S(D',D),T(D(x),B(y))),$$
$$B''_{\textmd{SBR}}(y)=\mathop{\bigvee}\limits_{x\in U}I(S(D',D),I(D(x),B(y))).$$

The following triple implication principle (TIP) for the GMP problem is proposed by Wang \cite{Wang1}.
\\{\bf Triple implication principle for GMP} Assume that the maximum of following formula
\begin{equation}M(x,y)=I(I(D(x),B(y)),
I(D'(x),B'(y)))
\end{equation}
exists for every $x\in U$ and $y\in V$, where $I$ is a fuzzy implication on [0,1]. The solution $B'$ of GMP problem should be the smallest fuzzy set on $V$ such that Eq.(1) achieves its maximum.
\\{\bf Lemma 2.23}\cite{Pei} i. If $I$ fulfills I2, then
$$\max_{x\in U,y\in V}M(x,y)=I(I(D(x),B(y)),
I(D'(x),1)),$$

ii. Moreover, if $I$ is right-continuous with respect to the second variable, then the TIP solution of GMP problem is unique.
\\{\bf Theorem 2.24}\cite{Pei} Let $I_T$ be an R-implication generated by a left-continuous t-norm $T$. Then the TIP solution of GMP problem is given by
$$B'_{\textmd{TIP}}(y)=\bigvee_{x\in U}T(D'(x),I_T(D(x),B(y))).$$
\section{ Satisfaction of LIA with fuzzy implications and aggregation functions}
\quad This section will study some properties of fuzzy implications and aggregation functions when they satisfy LIA.
\\{\bf Lemma 3.1} Let $I$ satisfy LIA with $A$. If $A$ is commutative, then $I$ satisfies EP.
\\{\bf Proof.} Straightforward.
\\{\bf Lemma 3.2} Let $I$ satisfy LIA with $A$. If $N_I(y_1)=N_I(y_2)$, then $N_I(A(x,y_1))=N_I(A(x,y_2))$ holds for arbitrary and fixed $x\in [0,1]$.
\\{\bf Proof.} Let $N_I(y_1)=N_I(y_2)$. We then have $N_I(A(x,y_1))=I(A(x,y_1),0)=I(x,I(y_1,0))=I(x,N_I(y_1))=I(x,N_I(y_2))=I(x,I(y_2,0))=I(A(x,y_2),0)=N_I(A(x,y_2))$.
\\{\bf Lemma 3.3} Let $I$ satisfy EP and $N_I$ be injective. If $I$ fulfills LIA with $A$, then $A$ is commutative.
\\{\bf Proof.} It suffices to take $z=0$ in LIA.
\\{\bf Lemma 3.4} Let $I$ be a fuzzy implication such that $I(x,y)=1$ iff $x=0$ or $y=1$. If $I$ satisfies LIA with $A$, then $A$ is
conjunctor.
\\{\bf Proof.} By LIA, we have $I(A(0,1),z)=I(0,I(1,z))=1$ for any $z\in[0,1)$. This implies $A(0,1)=0$. We can similarly obtain $A(1,0)=0$.
Thus, $A$ is a conjunctor.
\\{\bf Lemma 3.5} Let $I$ be a fuzzy implication such that $N_I$ is non-filling. If $I$ satisfies LIA with $A$, then $A$ is
conjunctor.
\\{\bf Proof.} By LIA, we have $N_I(A(0,1))=I(A(0,1),0)=I(0,I(1,0))=1$. This implies $A(0,1)=0$. We can similarly obtain $A(1,0)=0$.
Thus, $A$ is a conjunctor.
\\{\bf Lemma 3.6} Let the mapping  $h(z)=I(1,z)$ be continuous on [0, 1]. If $I$ satisfies LIA with $A$, then $I$ satisfies NP.
\\{\bf Proof.} For any $y\in [0,1]$, there exist some $z\in[0,1]$ such that $y=I(1,z)$ by the continuity of $h$.  Therefore, $I(1, y)=I(1,I(1,z))=I(A(1, 1), z)=I(1,z)=y$.
\\{\bf Definition 3.7}\cite{Liz} Let $I$ be a fuzzy implication  and $A$ an aggregation function. The pair $(I,A)$ is called an adjoint pair if
 they satisfy the residuation property (RP), i.e.
$$A(x,y)\leq z\Longleftrightarrow x\leq I(y,z), \ \forall x,y,z\in[0,1].$$
{\bf Lemma 3.8}  Let $I$ satisfy OP and LIA with $A$. We have

i.  $A$ is conjunctor,

ii. $(I,A)$ is an adjoint pair.
\\{\bf Proof.} i. By  LIA, we have $I(A(0,1),0)=I(0,I(1,0))=1$. Since $I$ fulfills OP, $A(0,1)=0$ holds. We can similarly obtain $A(1,0)=0$.
Thus, $A$ is a conjunctor.

 ii. Since $I$ satisfies OP, $I(x,I(y,z))=1\Longleftrightarrow x\leq I(y,z)$ holds for any $x,y,z\in[0,1]$. Similarly, $I(A(x,y),z)=1\Longleftrightarrow A(x,y)\leq z$. By LIA, we have $A(x,y)\leq z\Longleftrightarrow x\leq I(y,z)$.
\\{\bf Remark 1.} We can similarly obtain that $(I,A)$ forms an adjoint pair if $I$ satisfies OP$_\textmd{U}$ and LIA with $A$.
\\{\bf Lemma 3.9} Let $A$ be associative and commutative. If $I$ fulfills RP with $A$, then they satisfy LIA.
\\{\bf Proof.} By RP, $A(I(x, y), x))=A(x, I(x, y)))\leq y$ holds for any $x, y\in [0,1]$.  We can then assert that $I(x, I(y, z))\leq I(A(x,y),z)$. Indeed, $A(A(x,y),I(x, I(y, z)))=A(y,A(x,I(x, I(y, z))))$ $\leq z$. On the other hand, we have $A(A(x,y),I(A(x, y), z))\leq A(y, I(y, z))\leq z$. This means $A(x,I(A(x, y), z))\leq I(y,z)$. And then $I(A(x, y), z)\leq I(x,I(y,z))$.
\\{\bf Lemma 3.10} Let $A_\varphi$ and $I_\varphi$ be the $\varphi$-conjugate of $A$ and $I$, respectively. If $I$ fulfills LIA with $A$, then $I_\varphi$ satisfies LIA with $A_\varphi$.
\\{\bf Proof.} $I_\varphi(A_\varphi(x,y),z)=\varphi^{-1}(I(\varphi(A_\varphi(x,y)),\varphi(z)))=\varphi^{-1}(I(A(\varphi(x),\varphi (y)),\varphi(z)))=\varphi^{-1}$ $(I(\varphi(x),I(\varphi (y),\varphi(z))))=I_\varphi(x,\varphi^{-1}(I(\varphi (y),\varphi(z))))= I_\varphi(x ,I_\varphi(y,z))$.
\section{LIA with $(A,N)$- and R-implications}
\quad In this section, we shall seek some aggregation functions such that $(A, N)$- and R-implications satisfy LIA with them. We firstly consider the case when $(A, N)$-implications are generated by the smallest and greatest fuzzy negations, respectively.
\\{\bf Lemma 4.1} Let $I_{A,N}$ be an $(A, N)$-implication generated by an associative disjunctor $A$ and the smallest fuzzy negation $N_{\bot}$. Then, $I_{A,N}$ satisfies LIA with any conjunctor $A'$ without zero divisors.
\\{\bf Proof.} Let $A'$ be a conjunctor without zero divisors. We consider the following two cases.

i. $A'(x,y)=0$. This case implies $x=0$ or $y=0$. Then, we have $I_{A,N}(A'(x,y),z)=I_{A,N}(0,z)=1=I_{A,N}(x,I_{A,N}(y,z))$.

ii. $A'(x,y)\neq0$. In this case, we have $xy\neq0$. This implies
$I_{A,N}(A'(x,y),z)=A(0,z)$. On the other hand, $I_{A,N}(x,I_{A,N}(y,z))=A(0,A(0,z))=A(0,z)$. Therefore,  $I_{A,N}(A'(x,y),z)=I_{A,N}(x,I_{A,N}(y,z))$.
\\{\bf Lemma 4.2} Let $I_{A,N}$ be an $(A, N)$-implication generated by an associative disjunctor $A$ and the greatest fuzzy negation $N_{\top}$. Then, $I_{A,N}$ satisfies LIA with any conjunctor $A'$ without one divisors.
\\{\bf Proof.} This proof is similar to that of Lemma 4.1.

 However, it is not easy to seek some aggregation functions such that $(A,N)$-implications obtained from other non-continuous fuzzy negations satisfy LIA with them. So, we next focus on the $(A,N)$-implications generated by continuous fuzzy negations.
\\{\bf Theorem 4.3} Let $I_{A,N}$ be an $(A, N)$-implication generated by an associative disjunctor $A$ and a continuous fuzzy negation $N$. Then
 $I_{A,N}$ satisfies LIA with the aggregation function $A'$ defined by $A'(x,y)=\widetilde{N}(A(N(x),N(y))$.
\\{\bf Proof}. $I_{A,N}(A'(x,y),z)=A(N(A'(x, y)),z)=A(N(\widetilde{N}(A(N(x),N(y)))),z)=A(A(N(x),$ $N(y)),z)=A(N(x),A(N(y),z))=
A(N(x),I_{A,N}(y,z))=I_{A,N}(x,I_{A,N}(y,z))$.

 It is not difficult to see that other aggregation functions can be found such that the $I_{A,N}$ satisfies LIA with them, because there exists other fuzzy negation $N$ such that $N\circ\widetilde{N}=id$ holds. However, the following result shows that $A_N$ is the only one for $I_{A,N}$ satisfying LIA if $N_{I_{A,N}}$ is strict.
\\{\bf Theorem 4.4} Let $I_{A,N}$ be an $(A, N)$-implication generated by an associative disjunctor $A$ and a strict negation $N$.  If $N_{I_{A,N}}$ is an injective mapping, then $I_{A,N}$ satisfies LIA with $A'$ if and only if $A'$ is the $N$-dual of $A$.
\\{\bf Proof}. $(\Longleftarrow)$ This proof is similar to that of Theorem 4.3.

$(\Longrightarrow)$ Assume that $I_{A,N}$ satisfies LIA with $A'$, that is, $I_{A,N}(A'(x,y),z)$ $=I_{A,N}(x,I_{A,N}(y,z))$ holds for any $x,y,z\in[0,1]$. Setting $z=0$, we have $N_{I_{A,N}}(A'(x,y))=A(N(x),A(N(y),0))=A(A(N(x),N(y)),0)=N_{I_{A,N}}(N^{-1}(A(N(x),N(y))))$ for any $x,y\in[0,1]$. Since $N_{I_{A,N}}$ is injective, $A'(x,y))=N^{-1}(A(N(x),N(y)))$ holds for any $x,y\in[0,1]$. Therefore, $A'=A_N$.
\\{\bf Remark 3.} It is easy to see that $N_{I_{A,N}}=N$ holds if 0 is a right neutral element of $A$ and $N$ is strict. In this case, $I_{A,N}$ satisfies LIA with $A'$ iff $A'$ is the $N$-dual of $A$. Especially, an S-implication satisfies LIA with a t-norm $T$ iff $T$ is the $N$-dual of $S$.
\\{\bf Theorem 4.5} Let $I_{A,N}$ be an $(A, N)$-implication generated by a strict negation $N$.  Then $I_{A,N}$ satisfies LIA with  the $N$-dual of $A$ if and only if $A$ is associative.
\\{\bf Proof}. It is sufficient to verify that $A$ is associative. Since $I_{A,N}$ satisfies LIA with  the $N$-dual of $A$, we have $I_{A,N}(A_N(x,y),z)=A(A(N(x),N(y)),z)=A(N(x),A(N(y),z))=I_{A,N}(x,I_{A,N}(y,z))$. The continuity of $N$ implies that  $A$ is associative.

In the rest of this section, we study the law of importation for R-implications generated by an associative and commutative aggregation functions.  \\{\bf Theorem 4.6} Let $I_A$ be an R-implication generated by an associative, commutative and left-continuous aggregation function $A$.  We have

i. $I_{A}$ satisfies LIA with $A$.

ii. If $I_{A}$ fulfills OP, then $I_{A}$ satisfies LIA with $A'$ if and only if $A'=A$.
\\{\bf Proof}.  i. The proof comes from Lemma 3.1 in \cite{Liz} and Lemma 3.9.

ii. Let $I_{A}$ fulfill OP. Obviously, $I_A(A'(x,y),A'(x,y))=1$ holds for any $x,y\in[0,1]$. This implies $I_A(x,I_A(y,A'(x,y)))=1$ by LIA. Again, we obtain $x\leq I_A(y,A'(x,y))$. Thus, $A(x,y)\leq A'(x,y)$.

On the other hand,  $I_A(A'(x,y),A(x,y))=I_A(x,I_A(y,A(x,y)))\geq I_A(x,x)=1$ because $(I_A,A)$ is an adjoint pair. Then, we have $A'(x,y)\leq A(x,y)$.
\\{\bf Remark 4.} i. Indeed, $A(x,1)=A(1,x)=x$ holds for any $x\in [0,1]$ iff $I_{A}$ satisfies OP. This means that $I_{A}$ is an R-implication generated by the t-norm $T$. And then $I_{T}$ satisfies LIA with $A'$ iff $A'=T$. This result can be also found in Ref.\cite{Jayaram}.

 ii. Similarly, we can obtain the fact that $A$ is a uninorm if $I_{A}$ satisfies OP$_\textmd{U}$. Thus, $I_{A}$ is an R-implication generated by the uninorm $U$. And then $I_{U}$ satisfies LIA with $A'$ iff $A'=U$. This result can be also found in Ref.\cite{Massanet4}.

 iii. By Lemma 3.8, $(I_A,A')$ is an adjoint pair if $I_{A}$ satisfies OP (or OP$_\textmd{U}$) and LIA with $A'$. Theorem 4.6 shows that $A$ is a unique aggregation function such that $I_A$ satisfies RP with it in this case.

 iv. The R-implications generated by not left-continuous aggregation functions may not satisfy LIA with any aggregation function $A'$ or satisfy LIA with many aggregation functions as shown in the following examples.
 \\{\bf Example 4.7} Consider the Weber implication $I_{WB}$ defined as $I_{WB}(x,y)=\left\{\begin{array}{ll}
                                                                                        1 & x<1 \\
                                                                                        y & x=1
                                                                                     \end{array}\right.$.\vspace{1mm}
 Similar to the proof of Lemma 4.2, we can verify that $I_{WB}$ satisfies LIA with any conjunctor $A'$ without one divisors.\vspace{1mm}
 \\{\bf Example 4.8}  Let $A(x,y)=\left\{\begin{array}{cl}
                                                                                        0 & xy<0.5 \\
                                                                                        \frac{x+y}{2}&  \textmd{otherwise}
                                                                                     \end{array}\right.$.
It is obvious to see that $A$ is a not-left\vspace{2mm}-continuous conjunctor.  And then the R-implication $I_A$ generated by $A$ can be obtained as\vspace{1mm}
$I_A(x,y)=\left\{\begin{array}{cl}
                                                                                        1 & x<0.5 \\
                                                                                        \max(2y-x,\frac{1}{2x})&  \textmd{otherwise}
                                                                                     \end{array}\right.$.\vspace{1mm}

For any aggregation function $A'$, a simple calculation reveals $I_A(A'(1,1),0.8)=0.6$ and $I_A(1,I_A(1,0.8))=0.5$. This means that $I_A$ does not satisfy LIA with any aggregation function.
\section{Other implications satisfying LIA}
\quad \quad  In this section,
we investigate if the QL-, $f$-, $g$-, probabilistic, probabilistic S- and $T$-power implications satisfy LIA with some aggregation functions. Let $I_{A_1,A_2}$ be a QL-operation. Clearly, $I_{A_1,A_2}$ satisfies I3 and I5 when $A_1$ is a disjunctor and $A_2$ is a conjunctor. Therefore, we only consider the case where $I_{A_1,A_2}$ is obtained from a disjunctor, a conjunctor and a fuzzy negation in the rest of this section. Further, if $A_2$ has a right neutral element 1, then $I_{A_1,A_2}$ being a QL-implication implies that $A_1$ satisfies LEM. And then  the following statements hold.
\\{\bf Lemma 5.1} Let $I_{A_1,A_2}$ be a QL-implication generated by an associative disjunctor without one divisor $A_1$, a semi-copula $A_2$ and a fuzzy negation $N$. Then $I_{A_1,A_2}$ satisfies LIA with any aggregation function $A$ without zero divisors.
\\{\bf Proof.}  Since $A_1$ has not one divisors, it is not difficult to verify that $A_1$ satisfies LEM with respect to $N$ if and only if $N=N_\top$.  This implies that $I_{A_1,A_2}$ becomes an $(A,N)$-implication generated by an associative disjunctor $A_1$ and the greatest fuzzy negation $N_\top$. By Lemma 4.2, $I_{A_1,A_2}$ satisfies LIA with any aggregation function $A$ without zero divisors.
\\{\bf Theorem 5.2} Let $I_{A_1,A_2}$ be a QL-implication generated by a disjunctor $A_1$ such that the mapping $h(x)=A_1(x,0)$ is continuous on [0,1], a conjunctor $A_2$ and a continuous fuzzy negation $N$. If the  aggregation function $A$ is commutative, then $I_{A_1,A_2}$ satisfies LIA with $A$ if and only if $A(x,y)=\widetilde{N}_{I_{A_1,A_2}}
(I_{A_1,A_2}(\widetilde{N}_{I_{A_1,A_2}}(N_{I_{A_1,A_2}}(x)),N_{I_{A_1,A_2}}(y)))$, where $\widetilde{N}_{I_{A_1,A_2}}(x)=\vspace{1mm}\left\{\begin{array}{ll}
 N_{I_{A_1,A_2}}^{(-1)}(x), & x\in(0,1]\\
    1 & x = 0
    \end{array}\right.$.\vspace{1mm}
\\{\bf Proof.}  Since $A$ is commutative,  $I_{A_1,A_2}$ satisfies EP according to Lemma 3.1.   The continuity of $h(x)$ implies that $N_{I_{A_1,A_2}}(x)=A_1(N(x),0)$ is continuous. Therefore, $I_{A_1,A_2}$ can be rewritten as an $(S,N)$-implication generated by a t-conorm $S$ and the natural negation $N_{I_{A_1,A_2}}$ according to Theorem 2.4.10 in\cite{Baczynski}, where $S(x,y)=I_{A_1,A_2}(\widetilde{N}_{I_{A_1,A_2}}(x),y)$. By Theorem 4.3,
$I_{A_1,A_2}$ satisfies LIA with $A$ if and only if $A(x,y)=\widetilde{N}_{I_{A_1,A_2}}(S(N_{I_{A_1,A_2}}(x),N_{I_{A_1,A_2}}(y)))=\widetilde{N}_{I_{A_1,A_2}}
(I_{A_1,A_2}(\widetilde{N}_{I_{A_1,A_2}}(N_{I_{A_1,A_2}}(x)),N_{I_{A_1,A_2}}(y)))$.
\\{\bf Theorem 5.3} Let $I_f$ be an $f$-implication. $I_f$ satisfies LIA with an aggregation function $A$ if and only if $A(x,y)=xy$.
\\{\bf Proof}. $(\Longleftarrow)$ This can be verified directly.

 $(\Longrightarrow)$ By Theorem 2.14, $I_f$ can be rewritten as an $A$-implication $I_{A,N}$ generated by the disjunctor $A(x,y)=f^{-1}((1-x)f(y))$. Suppose that $I_f$ satisfies LIA with an aggregation function $A'$. Then, $A$ is associative according to Corollary 4.5. This means that $f(1-(1-x)(1-y))=(1-x)f(y)$ holds for any $x,y\in[0,1]$. Therefore, we obtain $A'(x,y)=A_N(x,y)=1-A(1-x,1-y)=xy$. That is,  $I_f$ satisfies LIA with an aggregation function $A'$ if and only if $A'(x,y)=xy$.
\\{\bf Lemma 5.4} Let $I_g$ be a $g$-implication. If $I_g$ satisfies LIA with an aggregation function $A$, then  $A$ has not zero divisors.
\\{\bf Proof.} On the contrary, we assume that there exist $x,y\in(0,1]$ such that $A(x,y)=0$. Since $I_g$ satisfies LIA with $A$, we have $I_g(A(x,y),0)=1$. However, $I_g(x,I_g(y,0))=I_g(x,0)=0$ by Proposition 3.2.7 in \cite{Baczynski}. This is a contradiction.
\\{\bf Theorem 5.5} Let $I_g$ be a $g$-implication. $I_g$ satisfies LIA with an aggregation function $A'$ if and only if $A'(x,y)=xy$.
\\{\bf Proof}. $(\Longleftarrow)$ This can be verified directly.

 $(\Longrightarrow)$ By Theorem 2.14, $I_g$ can be rewritten as an $A$-implication $I_{A,N}$ generated by the disjunctor $A(x,y)=g^{-1}(\frac{g(y)}{1-x}\wedge g(1))$. Assume that $I_g$ satisfies LIA with an aggregation function $A'$. Then, $A$ is associative and has not zero-divisors by Corollary 4.5 and Lemma 5.5. This implies that $A(x,A(y,z))=1$ iff $A(x,A(y,z))=1$. And then $\frac{g(z)}{(1-x)(1-y)}=\frac{g(z)}{1-g^{-1}(\frac{g(y)}{1-x})}$ holds for any $x,y,z\in[0,1)$. Thus, $A'(x,y)=A_N(x,y)=1-A(1-x,1-y)=xy$. That is,  $I_g$ satisfies LIA with an aggregation function $A'$ if and only if $A'(x,y)=xy$.
\\{\bf Remark 5.} Theorems 5.3 and 5.5 also appeared in \cite{Jayaram, Baczynski}, respectively. However, we provide two distinct proofs. And our proofs can help to understand them from another perspective.
\\{\bf Theorem 5.6} Let $I_C$ be a probabilistic implication. If the equation $x^2C\left(1-\frac{C(x,y)}{x},z\right)=xC\left(x,\frac{C(1-y,z)}{1-y}\right)-C(x,y)C\left(x,\frac{C(1-y,z)}{1-y}\right)$ holds for any $x,y,z\in[0,1]$ with understanding $\frac{0}{0}=1$, then $I_C$ satisfies LIA with an aggregation function $A'$ if and only if $A'(x,y)=\left\{\begin{array}{ll}
                                                                                                                      1-\frac{C(x,1-y)}{x}& x\neq0\\
                                                                                                                      0& x=0
                                                                                                                    \end{array}\right.$.\vspace{1mm}
 \\{\bf Proof.} By Theorem 2.14, $I_C$ can be rewritten as an $A$-implication $I_{A,N}$ generated by the\vspace{1mm} disjunctor $A(x,y)=\left\{\begin{array}{ll}
                                                                                                                      \frac{C(1-x,y)}{1-x}& x\neq1\\
                                                                                                                      1& x=1
                                                                                                                    \end{array}\right.$.
 The equation $x^2C\left(1-\frac{C(x,y)}{x},z\right)=xC\left(x,\frac{C(1-y,z)}{1-y}\right)-C(x,y)C\left(x,\frac{C(1-y,z)}{1-y}\right)$ can ensure that the disjunctor $A$ is associative. According to Theorem 4.4, $I_C$ satisfies LIA with an aggregation function $A'$ if and only if $A'$ is the $N$-dual of $A$. That\vspace{1mm} is, $A'(x,y)=\left\{\begin{array}{ll}
                                                                                                                      1-\frac{C(x,1-y)}{x}& x\neq0\\
                                                                                                                      0& x=0
                                                                                                                    \end{array}\right.$.\vspace{1mm}
 \\{\bf Remark 6.} Notice that there exist some  probabilistic implications which satisfy LIA with an aggregation function $A'$ without the condition of Theorem 5.6 (See Example 5.2 in \cite{Helbin}).
 \\{\bf Theorem 5.7} Let $\widetilde{I}_C$ be a probabilistic S-implication. If the equation $C(x,C(1-y,z)+y)=C(x,y)+C(x-C(x,y),z)$ holds for any $x,y,z\in[0,1]$, then  $\widetilde{I}_C$ satisfies LIA with an aggregation function $A'$ if and only if $A'(x,y)=x-C(x,1-y)$.
 \\{\bf Proof.} By Theorem 2.14, $\widetilde{I}_C$ can be rewritten as an $A$-implication $I_{A,N}$ generated by the disjunctor $A(x,y)=x+C(1-x,y)$.
 The equation $C(x,C(1-y,z)+y)=C(x,y)+C(x-C(x,y),z)$ implies that the disjunctor $A$ is associative. According to Theorem 4.4, $\widetilde{I}_C$ satisfies LIA with an aggregation function $A'$ if and only if $A'$ is the $N$-dual of $A$. That is, $A'(x,y)=x-C(x,1-y)$.
 \\{\bf Theorem 5.8} Let $T$ be a nilpotent t-norm with additive generator $t$. Then, its power implication $I^{T}$ does not satisfy LIA with any aggregation function.
\\{\bf Proof.} Suppose that $I^T$ satisfies LIA with an aggregation function $A$, that is, $I^T(A(x,y),z)=I^T(x,I^T(y,z))$. Taking $z=0$, we have
 $\frac{t(A(x,y))}{t(0)}=\frac{t(x)}{t\left(\frac{t(y)}{t(0)}\right)}\wedge1$. This means that $A$ is formed as $A(x,y)=t^{-1}\left(\frac{t(0)t(x)}{t\left(\frac{t(y)}{t(0)}\right)}\wedge t(0)\right)$.
 This case implies that 1 is a right neutral element of $A$. And\vspace{1mm} then $I^T(A(x,1),z)=\frac{t(x)}{t(z)}$ holds if $1>x>z$. However, $I^T(x,I^T(1,z))=I^T(x,0)=\frac{t(x)}{t(0)}$. Thus, $I^{T}$ does not satisfy LIA with any aggregation function.
 \\{\bf Theorem 5.9} Let $T$ be the minimum t-norm and a strict t-norm, respectively. Then, their power implications does not satisfy LIA with any aggregation function having zero divisors or being commutative.
 \\{\bf Proof.} We only consider the case where $T$ is minimum t-norm. Another case can be similarly proved. Suppose that the aggregation function $A$ has zero divisors and $A(x,y)=0$. Then, we have $I^{T_\textmd{M}}(A(x,y),0)=1$. However, $I^{T_\textmd{M}}(x,I^{T_\textmd{M}}(y,0))=0$.

 Moreover, we assume that $I^{T_\textmd{M}}$ satisfies LIA with a commutative aggregation function $A$. By Lemma 3.1, $I^{T_\textmd{M}}$ satisfies EP. However, $I^{T_\textmd{M}}$ does not satisfy EP (See Proposition 13 in \cite{Massanet}).
 \\{\bf Remark 7.} We can similarly verify that the $T$-power implications do not satisfy LIA with any aggregation function having a neutral element $e$, too. However, we cannot ensure whether they do not satisfy LIA with any aggregation function.
\section{LIA with a given \wzr{associative} aggregation function}
\qquad  For a fixed \wzr{associative} aggregation function $A$, this section aims to characterize the fuzzy implications satisfying LIA with $A$. We firstly extend Definition 6 in\cite{Massanet5} as follows.
\\{\bf Definition 6.1} Let $A$ be an aggregation function and $N$ a fuzzy negation. We say that $N$ is $A$-compatible if $N(y_1)=N(y_2)$ implies $N(A(x, y_1)) = N(A(x, y_2))$ for any $x\in [0,1]$.
\\{\bf Lemma 6.2} Let $I$ be a fuzzy implication and $N_I$ a continuous fuzzy negation. If $I$ satisfies LIA with a given conjunctor $A$, then
$I$ has the form of $I(x,y)=N_I(A(x,\widetilde{N}_I(y))$.
\\{\bf Proof.} Since $I$ satisfies LIA with a given conjunctor $A$, $I(A(x,y),z)=I(x,I(y,z))$ holds for any $x,y,z\in[0,1]$. Taking $z=0$, we have
 $N_I(A(x,y))=I(x,N_I(y))$. Let us consider the following two options.

 i. $y\in \textmd{Ran}(\widetilde{N}_I)$. In this case, we have $\widetilde{N}_I(N_I(y))=y$. This implies that  $I(x,N_I(y))=N_I(A(x,\widetilde{N}_I(N_I(y))))$. Since $N_I$ is continuous, we have $I(x,y)=N_I(A(x,\widetilde{N}_I(y))$.

 ii. $y\notin \textmd{Ran}(\widetilde{N}_I)$. This case means that there exists $y'\in\textmd{Ran}(\widetilde{N}_I)$ such that $N_I(y)=N_I(y')$. By Lemma 3.2, we have $N_I(A(x,y))=N_I(A(x,y'))$.  Therefore,  $I(x,N_I(y))=I(x,N_I(y'))=N_I(A(x,\widetilde{N}_I(N_I(y'))))=N_I(A(x,\widetilde{N}_I(N_I(y))))$ holds.  Since $N_I$ is continuous, we obtain $I(x,y)=N_I(A(x,\widetilde{N}_I(y))$.

 Further, considering $A$ is a conjunctor, it can be verified that $I(x,y)=N_I(A(x,\widetilde{N}_I(y))$ is a fuzzy implication.
\\{\bf Lemma 6.3} Let $A$ be an associative conjunctor and $N$ an $A$-compatible continuous fuzzy negation. Then $I(x,y)=N(A(x,\widetilde{N}(y))$ satisfies LIA with $A$.
\\{\bf Proof.} $I(A(x,y),z)=N(A(A(x,y),\widetilde{N}(z)))=N(A(x,A(y,\widetilde{N}(z)))$. Let us  consider the following two cases.

i. $A(y,\widetilde{N}(z))\in \textmd{Ran}(\widetilde{N})$. In this case, $N(A(x,A(y,\widetilde{N}(z))))
=N(A(x,\widetilde{N}(N(A(y,\widetilde{N}(z))))))=N(A(x,\widetilde{N}(I(y,z))))=I(x,I(y,z))$.

ii.  $A(y,\widetilde{N}(z))\notin \textmd{Ran}(\widetilde{N})$. This case implies that there exists
$y'\in\textmd{Ran}(\widetilde{N})$ such that $N(A(y,\widetilde{N}(z)))=N(y')$. Since $N$ is $A$-compatible, $N(A(x,A(y,\widetilde{N}(z))))=N(A(x,y'))$ holds for any $x\in[0,1]$. Therefore, $I(x,I(y,z))=N(A(x,\widetilde{N}(N(A(y,\widetilde{N}(z))))))=
N(A(x,\widetilde{N}(N(y'))))=N(A(x,y'))=I(A(x,y),z)$.
\\{\bf Theorem 6.4} Let $I$ be a fuzzy implication and $A$ a conjunctor. If $A$ is associative and $N_I$ is continuous, then
$I$ satisfies LIA with $A$ if and only if  $N_I$ is $A$-compatible and $I(x,y)=N_I(A(x,\widetilde{N}_I(y))$.
\\{\bf Proof.} $(\Longleftarrow)$ The proof comes from Lemma 6.3.

$(\Longrightarrow)$ Assume that $I$ satisfies LIA with $A$. By Lemma 3.2, $N_I$ is $A$-compatible. And then $I(x,y)=N_I(A(x,\widetilde{N}_I(y))$ according to Lemma 6.2.

In the rest of this section, we will characterize fuzzy implications for some distinguished conjunctors such that they satisfy LIA.
\\{\bf Lemma 6.5} Let $I$  be a fuzzy implication. Then $I$ satisfies LIA\vspace{1mm} with $C_\bot$ if and only if $I$ is the greatest fuzzy implication, that is, $I(x, y)=\left\{\begin{array}{ll}
                                                                  0& x=1, y=0\\
                                                                  1&\textmd{otherwise}
                                                                \end{array}\right.$.\vspace{1mm}
\\{\bf Proof.} $(\Longrightarrow)$ Assume that $I$ satisfies LIA with $C_\bot$. Then, $I(C_\bot(x,y),z)=I(x,I(y,z))$ holds for any $x,y,z\in[0,1]$. For any $x\neq1$, we have $I(x,0)=I(x,I(1,0))=I(C_\bot(x,1),0)=\vspace{1mm}$ $I(1,1)=1$. This implies that $I$ can be written as $I(x,y)=\left\{\begin{array}{ll}
                                                                  0& x=1,y=0\\
                                                                  1&\textmd{otherwise}
                                                                \end{array}\right.$.\vspace{1mm}

   $(\Longleftarrow)$ It is easy to verify that $I$ satisfies LIA with $C_\bot$.
\\{\bf Lemma 6.6} Let $I$  be a fuzzy implication and $N_I$ a continuous fuzzy negation. Then\vspace{1mm} $I$ satisfies LIA with $(C_{avg})_\top$ if and only if $N_I$ is non-filling and $I(x, y)=\left\{\begin{array}{ll}
                                                                  1& x=0 \ \textmd{or} \ y=1\\
                                                                  N_I(x)& y=0\\
                                                                  N_I(x)\wedge y&\textmd{otherwise}
                                                                \end{array}\right.$.\vspace{1mm}
\\{\bf Proof.}  $(\Longrightarrow)$ Suppose that $I$ satisfies LIA with $(C_{avg})_\top$. It can firstly be asserted that $N_I$ is a non-filling fuzzy negation. Otherwise, there exists $y_0\neq 0$ such that $N_I(y_0)=1$. Notice that $N_I(0)=N_I(y_0)=1$. This implies that $N_I((C_{avg})_\top(1,0))=N_I(0)=1>N_I((C_{avg})_\top(1,y_0))=N_I(1)=0$.

According to Lemma 6.2, we have\vspace{1mm}
 $I(x, y)=\left\{\begin{array}{ll}
                                                                   1& x=0 \ \textmd{or} \ y=1\\
                                                                  N_I(x)& y=0\\
                                                                  N_I(x)\wedge y&\textmd{otherwise}
                                                                \end{array}\right.$.\vspace{1mm}

   $(\Longleftarrow)$ Obviously, $(C_{avg})_\top$ is associative. Further, we can ensure that $N_I$ is $(C_{avg})_\top$-compatible. Without loss of  generality, we suppose that $N_I(y_1)=N_I(y_2)<1$ holds for $0<y_1<y_2$. In order to obtain $N_I((C_{avg})_\top(x,y_1))=N_I((C_{avg})_\top(x,y_2))$ for any $x\in[0,1]$, let us consider the following three options:

   i. $x\leq y_1<y_2$. In this case, we have $N_I((C_{avg})_\top(x,y_1))=N_I(y_1)=N_I((C_{avg})_\top(x,y_2))=N_I(y_2)$.

   ii. $y_1<x \leq y_2$. This case implies  $N_I(y_1)=N_I(x)=N_I(y_2)$. Therefore, we obtain $N_I((C_{avg})_\top(x,y_1))=N_I(x)=N_I((C_{avg})_\top(x,y_2))=N_I(y_2)$.

   iii. $y_1<y_2<x$. In this case, we have $N_I((C_{avg})_\top(x,y_1))=N_I(x)=N_I((C_{avg})_\top(x,y_2))$.

   Based on the argument above, $I$ satisfies LIA with $(C_{avg})_\top$ according to Lemma 6.3.

 By Theorem 3.7 in \cite{Klir}, $N$ is a fuzzy negation if and only if there exists a continuous strictly increasing function $g:[0, 1]\rightarrow [0, +\infty]$ with $g(0)=0$ such that $N(x)= g^{-1}(g(1)-g(x))$ for any $x\in [0,1]$. In this case, the representable aggregation function can be rewritten as $A(x,y)= g^{-1}((g(x)+g(y)-g(1))\vee 0)$. Then, we have the following result.
\\{\bf Lemma 6.7} Let $I$  be a fuzzy implication and $N_I$ a strict fuzzy negation. $I$ satisfies LIA with the representable aggregation function defined as $A(x,y)= g^{-1}((g(x)+g(y)-g(1))\vee 0)$ if and\vspace{1.5mm} only if $I(x, y)=\left\{\begin{array}{ll}
                                                                  1& f(N_I(x))+f(y)\leq f(0)\\
                                                                  f^{-1}(f(N_I(x))+f(y))&\textmd{otherwise}
                                                                \end{array}\right.$ with $f=g\circ N_I^{-1}$. \vspace{2mm}
 \\{\bf Proof.} $(\Longrightarrow)$ Suppose that $I$ satisfies LIA with $A$. By Lemma 6.2, we have\vspace{1mm}
 $I(x, y)=N_I(A(x,N_I^{-1}(y)))=N_I(g^{-1}((g(x)+g(N_I^{-1}(y))-g(1))\vee 0))=f^{-1}((f(N_I(x))+f(y)-f(0))\vee$\vspace{2mm}
$ 0)=\left\{\begin{array}{ll}
                                                                  1& f(N_I(x))+f(y)\leq f(0)\\
                                                                  f^{-1}(f(N_I(x))+f(y))&\textmd{otherwise}
                                                                \end{array}\right.$.\vspace{1.5mm}

   $(\Longleftarrow)$ This proof comes from Lemma 6.3.

 However, Lemma 6.2 cannot be used to characterize the fuzzy implications for $M_{\lambda,f}$ and $TS_{\lambda,f}$ by the aforementioned method because they are not associative.
\section{Fuzzy hierarchical inference engine with fuzzy implications satisfying LIA}
\qquad In this section, we will present three fuzzy hierarchical inference engines in MISO fuzzy systems based on the fuzzy implications satisfying LIA. Therefore, we assume that the fuzzy implication $I$ satisfies LIA with an aggregation function $A$ in this section. Firstly, let us study the solution of GMP problem in Pedrycz's,  Raha's and TIP methods, respectively.
\subsection{Three fuzzy hierarchical inference engines with fuzzy implication satisfying LIA}
{\bf Lemma 7.1} The solution of GMP problem in Pedrycz's method can be rewritten as
$$B'_{\textmd{BKS}}(y)=I(\bigvee_{x\in U}A(D'(x),D(x)),B(y)).$$
{\bf Proof.} $B'_{\textmd{BKS}}(y)=\mathop{\bigwedge}\limits_{x\in U} I(D'(x),I(D(x),B(y)))=\mathop{\bigwedge}\limits_{x\in U} I(A(D'(x),D(x)),B(y))
=I(\mathop{\bigvee}\limits_{x\in U}A$ $(D'(x),D(x)),B(y))$.
\\{\bf Lemma 7.2} The conclusion $B''_{\textmd{SBR}}$ of GMP problem in Raha's method  is
 $$B''_{\textmd{SBR}}(y)=\bigvee_{x\in U}I(A(S(D',D),D(x)),B(y)).$$
{\bf Proof.} Obvious.
\\{\bf Lemma 7.3} Let $I$ be a fuzzy
implication which is right continuous with respect to the second variable and satisfies OP. Then the TIP solution of GMP problem is
$$B'_{\textmd{TIP}}(y)=\bigvee_{x\in U}A(I(D(x),B(y)),D'(x)).$$
{\bf Proof.}  Since $I$ is right-continuous with respect to the second variable, the TIP solution of GMP problem is unique and  Eq.(1) takes its maximum 1 by Lemma 2.23. It is not difficult to verify that $I(I(D(x),B(y)),
I(D'(x),\mathop{\bigvee}\limits_{x\in U}A(I(D(x),B(y)),D'(x)))\equiv 1$ holds for any $x\in V$ and $y\in U$ according to Lemma 3.8.

On the other hand, assume that $C$ is an arbitrary fuzzy set on $V$ such that
$I(I(D(x),B(y)),$ $I(D'(x),C(y)))\equiv 1$ holds for any $x\in V$ and $y\in U$. Since $I$ satisfies LIA with the aggregation function $A$, we have
$I(I(D(x),B(y)),I(D'(x),C(y)))= I(A(I(D(x),B(y)), D'(x)),C(y))\equiv1$ for any $x\in V$ and $y\in U$.  The ordering property of $I$ implies that $C(y)\geq\mathop{\bigvee}\limits_{x\in U}A(I(D(x),B(y)),$ $D'(x))$. Therefore, $B'_{\textmd{TIP}}(y)=\mathop{\bigvee}\limits_{x\in U}A(I(D(x),B(y)),D'(x))$.\vspace{1mm}

In order to construct the fuzzy hierarchical inference engine in MISO fuzzy system, we combine the input and IF-THEN rules into the output by the above three methods. For convenience to show three fuzzy hierarchical inference engines, we only consider this case when $m=2$ and $n=1$ (that is, two-input-one-output fuzzy system and the fuzzy rule base including
only one rule).  Assume that
the fuzzifier is the singleton fuzzifier\cite{Wang} and that the aggregation function $A$  is employed to combine the antecedent of fuzzy IF-THEN rule in fuzzy inference engine. For an arbitrary input $\mathbf{x}_0=(x_{01}, x_{02})\in U_1\times U_2$, we have the following results.
\\{\bf Theorem 7.4} Let $I$ satisfy NP and $A$ be a conjunctor having a left neutral element 1. If the conjunctor $A$ is employed to combine the antecedent of IF-THEN rule and $I$ satisfies LIA, then the BKS inference engine is $B'_{\textmd{BKS}}=(D'_1,D'_2)\circ_{\textmd{BKS}}I((D_1,D_2),B)=D'_1\circ_{\textmd{BKS}}I(D_1,D'_2\circ_{\textmd{BKS}}I(D_2,B))$.
\\{\bf Proof.} By Lemma 7.1, $B'_{\textmd{BKS}}(y)=I(\mathop{\bigvee}\limits_{(x_1,x_2)\in U_1\times U_2}A(A(D'_1(x_1),D'_2(x_2)),A(D_1(x_1),D_2(x_2))),$\vspace{1mm} $B(y))=I(A(D_1(x_{01}),D_2(x_{02})),B(y))=I(D_1(x_{01}),I(D_2(x_{02}),B(y)))=
I(\mathop{\bigvee}\limits_{x_1\in U_1}A(D'_1(x_1),$\vspace{1mm} $D_1(x_1)),I(\mathop{\bigvee}\limits_{x_2\in U_2}A(D'_2(x_2),D_2(x_2)),B(y)))$. This can be shortened as
$B'_{\textmd{BKS}}=(D'_1,D'_2)\circ_{\textmd{BKS}}\vspace{1mm}I((D_1,D_2),B)=D'_1\circ_{\textmd{BKS}}I(D_1,D'_2\circ_{\textmd{BKS}}I(D_2,B))$.

For convenience, we shorten the conclusions $B''_{\textmd{SBR}}$ and $B'_{\textmd{TIP}}$ in Lemmas 7.2 and 7.3 as $B''_{\textmd{SBR}}=D'\circ_{\textmd{SBR}}I(D,B)$ and $B'_{\textmd{TIP}}=D'\circ_{\textmd{TIP}}I(D,B)$, respectively. Similar to Theorem 7.4, we obtain the following results.
\\{\bf Theorem 7.5} Let $A$ be an associative and commutative conjunctor and $S(A(D'_1,D'_2),A(D_1,$ $D_2))=A(S(D'_1,D_1),S(D'_2,D_2))$. If $A$ is employed to combine the antecedent of IF-THEN rule and $I$ satisfies LIA with $A$, then the SBR inference engine is $B''_{\textmd{SBR}}=(D'_1,D'_2)\circ_{\textmd{SBR}}I((D_1,D_2),B)=D'_1\circ_{\textmd{SBR}}I(D_1,D'_2\circ_{\textmd{SBR}}I(D_2,B))$.
\\{\bf Proof.} By Lemma 7.2, $B''_{\textmd{SBR}}(y)=\mathop{\bigvee}\limits_{(x_1,x_2)\in U_1\times U_2}I(A(S(A(D'_1,D'_2),A(D_1,D_2)), A(D_1(x_1),$\vspace{1mm} $D_2(x_2))),B(y))=I(A(S(A(D'_1,D'_2),A(D_1,D_2)), A(D_1(x_{01}),D_2(x_{02}))),B(y))=I(A(A(S(D'_1,$ $D_1),S(D'_2,D_2)), A(D_1(x_{01}),D_2(x_{02}))),B(y))=I(A(A(S(D'_1,D_1),D_1(x_{01})),A(S(D'_2,D_2),D_2$ $(x_{02}))),B(y))=I(A(S(D'_1,D_1),D_1(x_{01})),I(A(S(D'_2,D_2),D_2(x_{02})),B(y)))=
\mathop{\bigvee}\limits_{x_1\in U_1}I(A(S(D'_1,$\vspace{1mm} $D_1),D_1(x_1)),\mathop{\bigvee}\limits_{x_2\in U_2}I(A(S(D'_2,D_2), D_2(x_2)),B(y)))$. This can be shortened as
$B'_{\textmd{BKS}}=(D'_1,D'_2)$\vspace{1mm} $\circ_{\textmd{BKS}}I((D_1,D_2),B)=D'_1\circ_{\textmd{BKS}}I(D_1,D'_2\circ_{\textmd{BKS}}I(D_2,B))$.
\\{\bf Remark 8.} Since $D'_1$ and $D'_2$ are singleton fuzzy sets, the condition $S(A(D'_1,D'_2),A(D_1,D_2))=A(S(D'_1,D_1),S(D'_2,D_2))$ can be meet by some measures of similarity. For example, the several measures of similarity mentioned in Ref.\cite{Raha} satisfy this condition for any conjunctor.
\\{\bf Theorem 7.6} Let the conjunctor $A$ be employed to combine the antecedent of IF-THEN rule and $I$ satisfy LIA. If $I$ is right-continuous with respect to the second variable and fulfills OP, then the TIP inference engine is $B'_{\textmd{TIP}}=(D'_1,D'_2)\circ_{\textmd{TIP}}I((D_1,D_2),B)=D'_1\circ_{\textmd{TIP}}I(D_1,D'_2\circ_{\textmd{TIP}}I(D_2,B))$.
\\{\bf Proof.} According to Lemma 7.3, $B'_{\textmd{TIP}}(y)=\mathop{\bigvee}\limits_{(x_1,x_2)\in U_1\times U_2}A(I(A(D_1(x_1),D_2(x_2)),B(y)),A($\vspace{1mm} $D'_1(x_1),D'_2(x_2)))=I(A(D_1(x_{01}),D_2(x_{02})),B(y))=I(D_1(x_{01}),I(D_2(x_{02}),B(y)))=
\mathop{\bigvee}\limits_{x_1\in U_1}A(I$\vspace{1mm} $(D_1(x_1),\mathop{\bigvee}\limits_{x_2\in U_2}A(I(D_2(x_2)),B(y)),D'_2(x_2)),D'_1(x_1))$. This means
$B'_{\textmd{TIP}}=(D'_1,D'_2)\circ_{\textmd{TIP}}I((D_1,$\vspace{1mm} $D_2),B)=D'_1\circ_{\textmd{TIP}}I(D_1,D'_2\circ_{\textmd{TIP}}I(D_2,B))$.
\subsection{Discussion}
\qquad It is not difficult to see that these three hierarchical inference engines can be extended to any MISO fuzzy system. This implies that the MISO fuzzy system with these three inference engines can be converted into an SISO
hierarchical fuzzy system with these three inference engines. Similar to that mentioned in \cite{Jayaram}, it is sufficient to calculate the two-dimensional matrices
at each stage and to store the antecedent of fuzzy IF-THEN rules in the SISO
hierarchical fuzzy system. This means that the MISO fuzzy systems with these three inference engines have the advantages in storing and computing. However, in an $m$-input-one-output single
fuzzy system using the classical CRI method, Pedrycz's method and TIP method, when $|U_i|=n_i(i=1,2,\cdots, m)$
and $|V|=n$, the complexity of a single inference system
amounts to $O(n\prod_{i=1}^mn_i)$. Moreover, we have to calculate
an $m$-dimensional matrix having entries $\prod_{i=1}^mn_i$ if $|U_i|=n_i(i=1,2,\cdots, m)$. Therefore,
we need to store $m$-dimensional matrices for each fuzzy IF-THEN rule\cite{Cornelis,Stepnicka}.

Indeed, it owes to the law of importation that the MISO fuzzy system with these three inference engines is converted into a SISO
hierarchical fuzzy system. In the MISO fuzzy system, chosen the fuzzy implications (such as R-implication, $(A,N)$-implication, QL-implication and so on) to interpret the fuzzy IF-THEN rules in rule base, we can construct the aggregation functions such that they satisfy LIA by which obtained in Section 4. To enhance the storage and computational efficiency, people ought to accordingly employ these aggregation functions to combine the antecedent of fuzzy IF-THEN rules in rule base.

By the results in Section 5, if a given aggregation function is employed to translate the antecedent of fuzzy IF-THEN rules in rule base, we can also construct a fuzzy implication satisfying LIA. Similarly,  people should utilize the fuzzy implication to translate the fuzzy IF-THEN rules in rule base in order to advance the computational and storage efficiency.
\section{Numerical examples}
\qquad In this section, we will present three examples to illustrate the methods developed in the previous sections.
\\{\bf Example 8.1} Let $D_1=[0.9,0.7,0.9,0.6,0.8]$, $D_2=[1,0.7,0.8,0.9]$
and $B=[0.2,0.1,0.3]$ are fuzzy sets defined
on $U_1=\{x_{11},x_{12},x_{13},x_{14}\}$, $U_2=\{x_{21},x_{22},x_{23},x_{24}\}$ and
$V=\{y_1,y_2,y_3,$ $y_4, y_5\}$, respectively. We consider the two-input-one-output fuzzy system including the following single fuzzy IF-THEN rule

  \qquad\qquad\qquad\qquad\qquad IF \ $x_1$\ is\ $D_1$\ and\ $x_2$\ is\ $D_2$ THEN  \ $y$\ is \ $B$.

 Suppose that $I$ is the Kleene-Dienes implication $I_K(x,y)=(1-x)\vee y$. By Theorem 4.5, the Kleene-Dienes implication satisfies LIA with the minimum  t-norm $T_M$.
Now, we employ $T_M$ to combine the antecedent of IF-THEN rule. Let $D'_1=[0,1,0,0,0]$ and $D_2'=[0,0,1,0]$ be the fuzzy single input. In classical BKS inference method, we firstly compute $T_M(D_1',D_2')$ and  $T_M(D_1,D_2)$ as
$$T_M(D'_1,D'_2)=\left(
                 \begin{array}{cccc}
                   0& 0& 0& 0 \\
                   0& 0& 1& 0\\
                   0& 0& 0& 0 \\
                   0& 0& 0& 0 \\
                   0& 0& 0& 0 \\
                 \end{array}
               \right), \qquad T_M(D_1,D_2)=\left(
                 \begin{array}{cccc}
                   0.9 & 0.7 & 0.8& 0.9 \\
                   0.7& 0.7& 0.7 & 0.7\\
                    0.9 & 0.7 & 0.8& 0.9 \\
                   0.6& 0.6 & 0.6& 0.6 \\
                    0.8& 0.7 & 0.8& 0.8 \\
                 \end{array}
               \right).$$
Further, we have $\mathop{\bigvee}\limits_{(x_1,x_2)\in U_1\times U_2}T_M(T_M(D'_1(x_1),D'_2(x_2)),T_M(D_1(x_1),D_2(x_2)))=0.7$.\vspace{1mm} Therefore, the output is $B'_{\textmd{BKS}}=[0.3,0.3,0.3]$ \wzr{according to Lemma 7.1. The computational complexity to calculate the output $B'_{\textmd{BKS}}$ using the classical BKS method in this example can be considered as shown in Table 1.}
\wzr{$$\mbox{\bf{\small Table\ 1 \ The computational complexity of the classical BKS method}}$$}
\wzr{\begin{center}
\tabcolsep 0.05in
\begin{tabular}{ccc}
 \toprule[1pt]
  Stage & Process & Times\vspace{1mm}\\
\midrule[0.75pt]
 1& $T_M(D'_1(x_1),D'_2(x_2))$ &$5\times 4=20$\vspace{1mm}\\
  2& $T_M(D_1(x_1),D_2(x_2))$ &$5\times 4=20$\vspace{1mm}\\
 3& $T_M(T_M(D'_1(x_1),D'_2(x_2)),T_M(D_1(x_1),D_2(x_2)))$ &20\vspace{1.5mm}\\
4& $\mathop{\bigvee}\limits_{x_1\times x_2}T_M(T_M(D'_1(x_1),D'_2(x_2)),T_M(D_1(x_1),D_2(x_2)))$&19\\
 5& $B'_{\textmd{BKS}}$&3\vspace{1mm}\\
 Total & &82\vspace{1mm}\\
   \bottomrule[1pt]
\end{tabular}
\end{center}\vspace{1mm}}

\wzr{Moreover, some $5\times 4$-dimensional matrices are required to store for every fuzzy IF-THEN rule in order to compute the output.}

Next, we use the hierarchical BKS method proposed by Theorem 7.4 to compute the output $B'_{\textmd{BKS}}$. We have $T_M(D_2',D_2)=[1,0.7,0.8,0.9]$.
Then, $I_K(\mathop{\bigvee}\limits_{x_2\in U_2}T_M(D'_2(x_2),D_2(x_2),B(y))=[0.2,0.1,0.3]$. Further, we obtain $B'_{\textmd{BKS}}=I_K(\mathop{\bigvee}\limits_{x_1\in U_1}T_M(D'_1(x_1),D_1(x_1)),I_K(\mathop{\bigvee}\limits_{x_2\in U_2}T_M(D'_2(x_2),$ $D_2(x_2)),B(y)))=[0.3,0.3,0.3]$. Indeed, the result of the hierarchical BKS method proposed \wzr{by} Theorem 7.4 is equal to that of the classical BKS inference method. \wzr{Let us consider the computational complexity to calculate the output $B'_{\textmd{BKS}}$ using the hierarchical BKS method in this example. It is shown in Table 2.
$$\mbox{\bf{\small Table\ 2 \ The computational complexity of the hierarchical BKS method}}$$}
\wzr{\begin{center}
\tabcolsep 0.05in
\begin{tabular}{ccc}
 \toprule[1pt]
  Stage & Process & Times\vspace{1mm}\\
\midrule[0.75pt]
 1& $T_M(D'_2(x_2),D_2(x_2))$ &4\vspace{1mm}\\
  2& $\mathop{\bigvee}\limits_{x_2}T_M(D'_2(x_2),D_2(x_2))$ &3\vspace{1mm}\\
 3& $I_K(\mathop{\bigvee}\limits_{x_2\in U_2}T_M(D'_2(x_2),D_2(x_2),B(y))$ &3\vspace{1.5mm}\\
 4& $T_M(D'_1(x_1),D_1(x_1))$ &4\vspace{1mm}\\
5& $\mathop{\bigvee}\limits_{x_1}T_M(D'_1(x_1),D_1(x_1))$ &3\vspace{1.5mm}\\
 6& $B'_{\textmd{BKS}}$&3\vspace{1mm}\\
 Total & &20\vspace{1mm}\\
   \bottomrule[1pt]
\end{tabular}
\end{center}\vspace{1mm}}

\wzr{Moreover, we only need to store the different antecedent fuzzy sets (that is, some 3-dimensional vectors) for every fuzzy IF-THEN rule when the output is computed. Compared Table 1 with Table 2, it is obvious to see that the hierarchical BKS method helps to enhance the computational efficiency of the fuzzy inference engine.}
\\{\bf Example 8.2} Let the fuzzy sets $D_1$, $D_2$ and $B$ be as defined in
Example 8.1. We consider the same two-input-one-output fuzzy system.

Assume that the aggregation function is the greatest averaging conjunctor $(C_{avg})_\top$. We can find that the fuzzy implication $I(x,y)=\left\{\begin{array}{ll}
 1& x=0\ \textmd{or}\ y=1\\
1-x&y=0\\
(1-x)\wedge y& \textmd{otherwise}\\
\end{array}\right.$ satisfies LIA with $(C_{avg})_\top$ by Lemma 6.6.
 Thus, we utilize $(C_{avg})_\top$ to combine the antecedent of fuzzy IF-THEN rule.  And $S(A,B)=1-\mathop{\max}\limits_{x\in U}|A(x)-B(x)|$ is used to measure the similarity between two fuzzy sets $A$ and $B$. Let $D'_1=[0,1,0,0,0]$ and $D_2'=[0,0,1,0]$ be the fuzzy single input. In classical SBR inference method, we firstly compute $(C_{avg})_\top(D_1',D_2')$ and $(C_{avg})_\top(D_1,D_2)$ as follows
$$(C_{avg})_\top(D_1',D_2')=\left(
                 \begin{array}{cccc}
                   0 & 0 & 0& 0\\
                   0& 0& 1& 0\\
                   0& 0& 0& 0\\
                   0& 0& 0& 0\\
                    0& 0& 0& 0\\
                 \end{array}
               \right), \quad(C_{avg})_\top(D_1,D_2)=\left(
                 \begin{array}{cccc}
                   1 & 0.9 & 0.9& 0.9\\
                   1& 0.7& 0.8& 0.9\\
                   1 & 0.9 & 0.9& 0.9\\
                    1& 0.7& 0.8& 0.9\\
                    1& 0.8& 0.8& 0.9\\
                 \end{array}
               \right).$$
Further, we have $S((C_{avg})_\top(D_1',D_2'),(C_{avg})_\top(D_1,D_2))=0.8$. Then, $(C_{avg})_\top(S((C_{avg})_\top(D_1',$ $D_2'), (C_{avg})_\top(D_1,D_2)),(C_{avg})_\top(D_1,D_2))$ can be obtained as
$$(C_{avg})_\top(S((C_{avg})_\top(D_1',D_2'),(C_{avg})_\top(D_1,D_2)),(C_{avg})_\top(D_1,D_2))=\left(
                 \begin{array}{cccc}
                   1 & 0.9 & 0.9& 0.9\\
                   1& 0.8& 0.8& 0.9\\
                   1 & 0.9 & 0.9& 0.9\\
                    1& 0.8& 0.8& 0.9\\
                    1& 0.8& 0.8& 0.9\\
                 \end{array}
               \right).$$
Finally, we obtain the output $B''_{\textmd{SBR}}=[0.2,0.1,0.2]$ \wzr{by Lemma 7.2}.

Next, we calculate the output $B'_{\textmd{BKS}}$ with the hierarchical SBR method proposed by Theorem 7.5. Obviously, $S(D_2',D_2)=0.8$. Therefore,  we have $D_2'\circ_{\textmd{SBR}}I(D_2,B)=[0.2,0.1,0.2]$. After some tedious computations, we obtain $B''_{\textmd{SBR}}=D_1'\circ_{\textmd{SBR}}I(D_2'\circ_{\textmd{SBR}}I(D_2,B))=[0.2,0.1,0.2]$.
Obviously, the result of hierarchical SBR method is equal to that of the classical SBR inference method. \wzr{Indeed, we only need to calculate a $5\times 4$-dimensional matrix at every stage involved in the hierarchical SBR method. This improves the computational efficiency of fuzzy inference engine.}
\\{\bf Example 8.3} Let the fuzzy sets $D_1$, $D_2$ and $B$ be as in
Example 8.1, with the same structure for the given fuzzy IF-THEN
rule.

 Let $I$ be the {\L}ukasiewicz implication $I_L(x,y)=(1-x+y)\wedge 1$. According to Remark 4, the {\L}ukasiewicz implication satisfies LIA with {\L}ukasiewicz t-norm $T_L$, that is, $T_L(x,y)=(x+y-1)\vee 0$.
Thus, $T_L$ is used to combine the antecedent of fuzzy IF-THEN rule. In classical TIP inference method, we firstly compute the Cartesian product of $D_1$ and $D_2$ with respect
to $T_L$ as follows
$$T_L(D_1,D_2)=\left(
                 \begin{array}{cccc}
                   0.9 & 0.6 & 0.7& 0.8 \\
                   0.7& 0.4& 0.5 & 0.6\\
                   0.9& 0.6 & 0.7 & 0.8 \\
                   0.6& 0.3 & 0.5& 0.5 \\
                    0.8& 0.5 & 0.6& 0.7 \\
                 \end{array}
               \right).$$
Further, we have $I_L(T_L(D_1,D_2),B)=I_L(T_L(D_1,D_2), [0.2,0.1,0.3])$. Concretely,
$$I_L(T_L(D_1,D_2),B(y_1))=\left(
                 \begin{array}{cccc}
                   0.3 & 0.6 & 0.5& 0.4 \\
                   0.5& 0.8 & 0.7& 0.6\\
                   0.3& 0.6 & 0.5 & 0.4 \\
                   0.6& 0.9 & 0.7 & 0.7 \\
                    0.4& 0.6 & 0.6& 0.5 \\
                 \end{array}
               \right),$$ $$I_L(T_L(D_1,D_2),B(y_2))=\left(
                 \begin{array}{cccc}
                   0.2 & 0.5 & 0.4& 0.3 \\
                   0.4& 0.7& 0.6& 0.5\\
                   0.2& 0.5 & 0.4 & 0.3 \\
                   0.5& 0.8 & 0.6 & 0.6 \\
                    0.3& 0.5 & 0.5& 0.4 \\
                 \end{array}
               \right), $$
               $$I_L(T_L(D_1,D_2),B(y_3))=\left(
                 \begin{array}{cccc}
                   0.4 & 0.7 & 0.6& 0.5 \\
                   0.6& 0.9 & 0.8& 0.7\\
                   0.4& 0.7 & 0.6 & 0.5 \\
                   0.7& 1 & 0.8 & 0.8 \\
                    0.5& 0.7 & 0.7& 0.6 \\
                 \end{array}
               \right).$$
Let $D'_1=[0,1,0,0,0]$ and $D_2'=[0,0,1,0]$ be the fuzzy single input. \wzr{Then
$$T_L(D'_1,D'_2)=\left(
                 \begin{array}{cccc}
                   0 & 0 & 0& 0 \\
                   0& 0 & 1& 0\\
                   0& 0 & 0 & 0 \\
                   0& 0 & 0 & 0 \\
                    0& 0& 0& 0 \\
                 \end{array}
               \right).$$}
 By Lemma 7.3, we obtain the output $B'_{\textmd{TIP}}=\wzr{T_L(D'_1,D'_2)\circ_{\textmd{TIP}}I_L(T_L(D'_1,D'_2), B)}=[0.7,0.6,0.8]$.

Next, we apply the hierarchical TIP method proposed by Theorem 7.6 to compute the output $B'_{\textmd{TIP}}$. Given the input $(D_1',D_2')$, we have
$$I_L(D_2,B)=\left(
                 \begin{array}{ccc}
                   0.2 & 0.1 & 0.3 \\
                   0.5& 0.4 & 0.6\\
                   0.4& 0.3 & 0.5\\
                   0.3& 0.2 & 0.4 \\
                 \end{array}
               \right).$$
And then $D_2'\circ_{\textmd{TIP}}I_L(D_2,B)=[0.4,0.3,0.5]$. Further, $I_L(D_1, D_2'\circ_{\textmd{TIP}}I_L(D_2,B))$ can be calculated as follows
$$I_L(D_1, D_2'\circ_{\textmd{TIP}}I_L(D_2,B))=\left(
                 \begin{array}{ccc}
                   0.5 & 0.4 & 0.6 \\
                   0.7& 0.6 & 0.8\\
                   0.5 & 0.4 & 0.6 \\
                   0.8& 0.7 & 0.9\\
                   0.6& 0.5 & 0.7 \\
                 \end{array}
               \right).$$
Finally, we obtain the output $B'_{\textmd{TIP}}=D_2'\circ_{\textmd{TIP}}I_L(D_1, D_2'\circ_{\textmd{TIP}}I_L(D_2,B))=[0.7,0.6,0.8]$. Clearly, the result of the  hierarchical TIP method proposed Theorem 7.6 is equal to that of the classical TIP inference method.

\section{Conclusions}
\qquad We firstly have studied the fuzzy implications satisfying LIA. And then three hierarchical inference engines employed aggregation functions and fuzzy implications such that they satisfy LIA have been investigated. Specifically, we have

(1)  Analyzed the properties of aggregation functions and fuzzy implications when they satisfy LIA;

(2) Given the necessary and sufficient conditions for $(A,N)$-implications \wzr{generated by an associative disjunctor} and R-implications \wzr{generated by an associative and commutative aggregation function} satisfying LIA with some aggregation functions;

(3) Found some \wzr{associative} aggregation functions for $f$-implication, $g$-implication, QL-implication, probabilistic implication, probabilistic S-implication and $T$-power implications such that they satisfy LIA;

(4) Characterized the fuzzy implications satisfying LIA with a given \wzr{associative} aggregation function;

(5) Constructed three fuzzy hierarchical inference engines in MISO fuzzy system utilized the aggregation functions and fuzzy implications satisfying LIA.

(6) Demonstrated that our proposed methods are efficient.

Our results can help to improve the effectiveness of fuzzy inference engine in MISO fuzzy systems. In the future, we
wish to study the capability of fuzzy systems with these hierarchical inference engines. We also will apply them in real-life control problems and decision making.
\section{Acknowledgement}  \qquad This work was supported by the National Natural
Science Foundation of China (Grant No. 61673352).

\end{document}